\documentclass[11pt,a4paper]{article}
\usepackage{emnlp2018}
\usepackage{times}
\usepackage{latexsym}

\usepackage{url}

\aclfinalcopy

\usepackage[utf8]{inputenc}
\usepackage{verbatim}
\usepackage{booktabs}
\usepackage{multirow}

\newcommand{\eat}[1]{}

\newcommand{\veronica}[1]{\textcolor{magenta}{$_{V}$[#1]}}

\usepackage{colortbl}
\usepackage{pgfplots}
\definecolor{darkgreen}{RGB}{87,187,138}
\definecolor{darkgold}{RGB}{255,214,102}
\definecolor{darkred}{RGB}{230,124,115}
\newcommand{\ColorCell}[4]{%
        \newcommand*{\MinNumber}{#2}
        \newcommand*{\MaxNumber}{#3}
        \newcommand*{\MidNumber}{#4}
        \ifdim #1 pt >\MidNumber pt \relax
            \pgfmathsetmacro{\PercentColor}{max(min(100.0*(#1 - \MidNumber)/(\MaxNumber-\MidNumber),100.0),0.00)}
            \xdef\PercentColor{\PercentColor}
            \cellcolor{darkgreen!\PercentColor!darkgold}{#1}
        \else
            \pgfmathsetmacro{\PercentColor}{max(min(100.0*(\MidNumber - #1)/(\MidNumber-\MinNumber),100.0),0.00)}
            \xdef\PercentColor{\PercentColor}
            \cellcolor{darkred!\PercentColor!darkgold}{#1}
        \fi
}

\title{Residualized Factor Adaptation\\for Community Social Media Prediction Tasks}

\author{Mohammadzaman Zamani,\textsuperscript{1} H. Andrew Schwartz,\textsuperscript{1} Veronica E. Lynn,\textsuperscript{1}  \\ {\bf Salvatore Giorgi,\textsuperscript{2}} and {\bf Niranjan Balasubramanian\textsuperscript{1}  }\\
  \textsuperscript{1} Computer Science Department, Stony Brook University\\
  \textsuperscript{2}Department of Psychology, University of Pennsylvania \\
  {\tt mzamani@cs.stonybrook.edu} \\
  }

\date{November 2017}

\begin{document}

\maketitle

\begin{abstract}
    Predictive models over social media language have shown promise in capturing community outcomes, but approaches thus far largely neglect the socio-demographic context (e.g.~age, education rates, race) of the community from which the language originates.   
    For example, it may be inaccurate to assume people in Mobile, Alabama, where the population is relatively older, will use words the same way as those from San Francisco, where the median age is younger with a higher rate of college education.
    In this paper, we present \textit{residualized factor adaptation}, a novel approach to community prediction tasks which both (a) effectively integrates community attributes, as well as (b) adapts linguistic features to community attributes (factors). 
    We use eleven demographic and socioeconomic attributes, and evaluate our approach over five different community-level predictive tasks, spanning health (heart disease mortality, percent fair/poor health), psychology (life satisfaction), and economics (percent housing price increase, foreclosure rate). 
    Our evaluation shows that \textit{residualized factor adaptation}  significantly improves 4 out of 5 community-level outcome predictions over prior state-of-the-art for incorporating socio-demographic contexts.
\end{abstract}

\section{Introduction}
Adapting to human factors has been shown to benefit NLP tasks, especially in tasks that involve predictions over individual social media posts (e.g., sentiment~
\cite{hovy2015demographic}, sarcasm, and stance detection~\cite{lynn2017human}).
The main idea behind these approaches is that knowing who wrote a piece of text can help models better understand how to process it. 
This paper develops methods that apply this idea to community-level prediction tasks, which require making decisions over posts from a community of users. Many community-level outcomes and community-wide language are linked to socio-demographic factors (age, gender, race, education, income levels) with many social scientific studies supporting their predictive value~\cite{cohen2003poverty}, and should therefore affect how a model treats social media-based language features. For example, a high prevalence of the word ``bike'' in San Francisco, CA might be a signal that exercise is common in the area, while its high prevalence in Mobile, Alabama might  indicate greater interest in motor bikes. We present a method for building language-based predictive models which integrate in and adapt to attributes of the communities generating the language.

This work aims to unify two different approaches developed for adapting to human factors and use them for incorporating community attributes in community-level prediction tasks: 
(1) \textit{residualized controls:} whereby a model is trained in two steps: first over the factors/controls and then fitting the language to the residuals of the control model~\cite{zamani2017using}, and
(2) \textit{user-factor adaptation:} whereby linguistic features are adapted, or treated differently, based on the continuous-valued factors of the authors of the features~\cite{lynn2017human}.

Combining factor adaptation (FA) and residualized control (RC) into RFA is a non-trivial task. The intent behind both methods are quite different: whereas RC attempts to address the inherent heterogeneity between robust control variables and noisy linguistic variables, FA enables a model to treat linguistic features differently depending on the factors. From a statistical learning perspective, RC separates inference over controls from inference over language (model level integration), while FA brings controls and language together and makes the inference as one single step (data level integration). 
Additionally, FA has stricter bounds in the number of factors it can accommodate because each new factor has a multiplicative effect on the number of learned parameters. 
On the other hand, each new factor for RC typically only adds one new parameter. 
Here, we endeavour to develop RFA such that it achieves the benefits of both approaches with little lost to the limitations.
RFA inherits the challenges of the FA method with feature explosion. We address this through a systematic exploration of both feature and factor selection.

The main contributions of this work include: (1) the introduction of residualized factor adaptation which effectively combines extra-linguistic and language features,
(2) the first empirical evaluation of applying factor adaptation for community-level prediction tasks, (3) analysis of the impact of the size of factors and factor selection in adaptation, and (4) state-of-the-art accuracies for each of the five tasks for which we evaluate RFA.

\eat{Contributions:

1. introduce residualized factor adaptation

2. adaptation for community-level prediction using community-level factors.

3. analyze the impact of factor selection in adaptation.
}

\section{Background}
Social media provides easy access to a vast amount of language written by a diverse group of users, making it an increasingly popular resource for measuring community health, psychology, and economics~\cite{coppersmith2015adhd,Eichstaedt2015psychological,weeg2015using,mowery2016towards,haimson2017changes}. \cite{coppersmith2015adhd}, for instance, examine trends in language use among Twitter users who self-reported one of ten mental health diagnoses. \cite{Eichstaedt2015psychological} and \cite{weeg2015using} use Twitter to predict the prevalence rates of various health outcomes, such as heart disease mortality and depression, at the county level. \cite{haimson2017changes} tracked changes in the emotional well-being of transgender communities on Tumblr between 2015 and 2016. 

Socio-demographics are often correlated with health outcomes (such as age and heart disease), which is why such variables are often used as controls during analysis~\cite{coppersmith2015adhd,dos2015using,Eichstaedt2015psychological,weeg2015using}. Because of their predictive power, socio-demographics and other extra-linguistic information can additionally be leveraged when building the model itself. 

However, a central challenge in integrating community attributes is that they have very different properties than linguistic features and can be lost, in essence, like a needle in a haystack. 
For example, linguistic features like n-grams are high dimensional, with each dimension having high covariance with other dimensions and likely very little relationship with the outcome. On the other hand community features may be measured more robustly and are relatively low dimensional, often obtained through well-defined measurements.  
Not surprisingly, \cite{zamani2017using} showed a naive combination that simply concatenates these two sets of features risks losing the effective extra-linguistic features in a sea of weak linguistic features. 
They go on to show a \textit{residualized control} approach achieves significantly greater accuracy at economic prediction by first learning a model using extra-linguistic features (i.e.~controls or community factors) and then train a language model on top of the residual error of the previous model. 

It is possible that even when extra-linguistic features are not directly beneficial for prediction, they can still affect people's language.
Other related works consider how the meaning of language changes depending on who states it.  
For instance, when an NLP PhD student says the word `paper' he/she usually means something different than when a $5^{th}$ grade student uses the same word (i.e.~`research paper' versus `piece of paper'). 
\cite{hu2017world} noted the same words can have different meanings if different people say them. 
This idea of contextualizing language with extra-linguistic information has been the basis for multiple models: \cite{hovy2015demographic} learn age- and gender-specific word embeddings, leading to significant improvements for three text classification tasks. \cite{Volkova2013EMNLP} found that using gender-specific features lead to improvements in sentiment analysis over a gender-agnostic model. Most recently, \cite{lynn2017human} proposed a domain adaptation-inspired method for composing user-level, extra-linguistic information with message-level features, leading to improvements for multiple text classification tasks; we build off of this approach and that of \cite{zamani2017using} in this paper. 

While \cite{lynn2017human} injected user-level info into message-level tasks, we are investigating whether same-level adaptation techniques are similarly useful.

We also try to find the circumstances under which each of the adaptation and residualized control approaches are more powerful, and we take on the non-trivial task of exploiting concepts from both the adaptation and the residualized control techniques at the same time, finding that they add even more power when combined with one another. 

\section{Method}

We describe \textit{residualized factor adaptation} (RFA), an approach to text-based prediction utilizing extra-linguistic factors (also called controls -- often demographic or socioeconomic information).
The key challenge for RFA lies in effectively combining two different types of features. The language-based features, extracted from the tweets, are numerous but are only weak indicators of the outcomes. The socioeconomic and demographic features, on the other hand, are strong indicators but fewer in number. 
Naively combining both sets of features ignores this crucial difference in their predictive abilities, potentially resulting in important features getting drowned out. 

We first describe two methods that effectively combine extra-linguistic factors at two different levels: 1) \textit{Residualized control} is a model-level combination method which builds different models for each type of feature, then combines the results of these models to make the final outcome prediction.
2) \textit{Factor adaptation} is a feature-level combination method that composes the two feature sets with one another to produce a transformed feature space over which a single model may be built.
Finally, we present our combined method of \textit{Residualized Factor Adaptation} which takes advantage of both concepts without exploding model parameters.

\subsection{Residualized Control Prediction}
Language-based features and community-level attributes are qualitatively different modalities. The extra-linguistic variables, while few in number, are mostly unbiased and follow a normal distribution, which can be used to build a strong outcome predictor. However, without special treatment, the signal in extra-linguistic variables can be overwhelmed when combined with a large number of language-based features. 

The residualized control approach~\cite{zamani2017using} avoids this issue by 
building two models. The first is a prediction model built over the extra-linguistic variables (or \textit{controls}) alone.
The error, or \textit{residuals}, produced by this first model represents the information that was unable to be predicted using the extra-linguistic variables alone. The language-based features are therefore brought in to improve upon the initial predictions by using the residuals as training labels for a model based on the linguistic features. In this way, the language-based features are able to account for additional information not captured by the initial extra-linguistic feature-only model. At test time, each instance is fed to both prediction models, and the final outcome is given as a sum of the predictions from both models --- the outcome predicted by the extra-linguistic model adjusted for error by the language-based model.

Formally, given extra-linguistic features $X_{EL}$ and language features $X_L$, the residualized control models are built as follows:
\begin{equation}
    \label{eq:control}
    \hat{Y} = \alpha \times X_{EL} + \beta 
    \vspace*{-6pt}
\end{equation}
\begin{equation}
    \label{eq:error}
    \epsilon = Y - \hat{Y}
    \vspace*{-6pt}
\end{equation}
\begin{equation}
    \label{eq:language}
    \epsilon \simeq \gamma \times X_{L} + \lambda
\end{equation}

The extra-linguistic control model is parameterized by $\alpha$ weights and the $\beta$ bias term. $\epsilon$ denotes the residual, i.e., the error of the extra-linguistic model. The language-based model aims to predict the $\epsilon$ residuals, with $\gamma$ weights and the $\lambda$ bias term as parameters. 

The motivation for this approach is that extra-linguistic features are more informative and less noisy than the language ones. By exploiting this two-stage learning procedure, the model is biased toward favoring the role of extra-linguistics over language features, which prevents the powerful but rare extra-linguistic features from being lost among thousands of noisy language features.

\subsection{Factor Adaptation}

\cite{lynn2017human} introduced \textit{user-factor adaptation}, a technique for combining message-level features with user-level information (or \textit{factors}) at the feature level. User-factor adaptation, which is based on the feature augmentation approach for domain adaptation~\cite{III2007ACL}, uses the extra-linguistic features to transform the language-based features. 
Each of the language-based features has additional, corresponding features that are a composite of itself and an extra-linguistic factor.
In this way, the model is able to capture both factor-specific and factor-general properties of each of the language-based features.

Following the work of \cite{lynn2017human}, we use a multiplicative composition function for combining the linguistic and extra-linguistic features. Instead of using user-level factors, we use extra-linguistic variables obtained at the community level, as described below. More formally, let $V_j$ be a matrix such that: 
\begin{equation}
    \forall j \in \{0,d\} : V_j = v_j \odot \mathbf{1}^l
    \label{eq:factor_mat}
\end{equation}

where $d$ is the number of extra-linguistic factors. With $n$ as the number of data instances, let $v_j$ be a column vector of height $n$ where element $v_{j,i}$ is the score of extra-linguistic feature $j$ for instance $i$. Having $l$ as the number of language variables, in Eq.~\ref{eq:factor_mat} for each factor $j$ we make a matrix of size $n \times l$, named $V_j$, in which every column is equal to $v_j$, and $V_j$ has the same dimensions as language feature matrix $X_L$. Now for each factor $j$ we use the Hadamard product to multiply $V_j$ with $X_L$. In this way each row of $X_L$ will be multiplied by the corresponding row in $V_j$, which is also equal to the corresponding value in $v_j$. 
We therefore can write the factor adaptation as follows:

\begin{equation}
    \label{eq:adaptation}
    X_{A} =  \lbrack V_{1} \odot X_L,V_{2} \odot X_L, \cdots, V_{d} \odot X_L \rbrack 
\end{equation}
The adapted features together with the original language-based features are used for building a single prediction model:
\begin{equation}
    \label{eq:adaptation-model}
    \hat{Y} = \alpha \times \lbrack X_{L} , X_{A} \rbrack + \beta 
\end{equation}

\subsection{Residualized Factor Adaptation}

Even though both the residualized control and factor adaptation approaches exploit extra-linguistics, they combine these in very different ways. The former does it at the model level by learning different models for different types of features and combining those models together. The latter does it at the data level by first combining both sets of features into a transformed feature set and then learning a single model on the obtained features. 
In addition, these approaches have different motivations and aim to accomplish different objectives.

\begin{figure}[t!]
    \centering
    \includegraphics[scale=0.25]{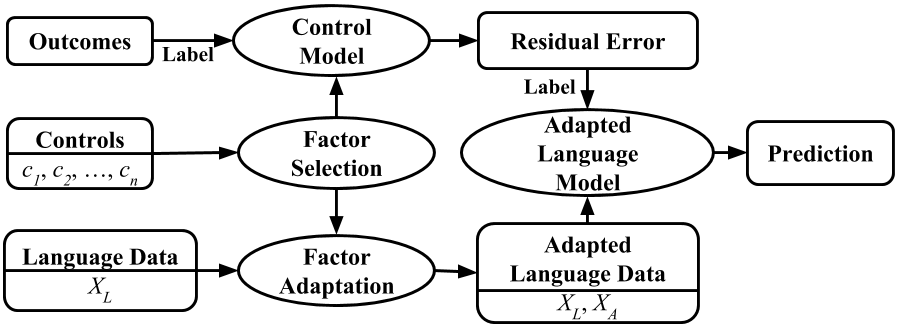}
    \caption{Components of residualized factor adaptation. $X_L$ is language data (topic and n-gram features) and $X_A$ is adapted language data.}
    \label{fig:rfa}
\end{figure}
These modeling differences suggest that the two approaches could have complementary benefits. 
Residualized factor adaptation (RFA), our proposed method, inherits the advantages of both the residualized control and adaptation techniques, and is depicted in Fig.~\ref{fig:rfa}. There are four main steps:

\noindent{\bf Step 1: Extra-linguistic control model.} We build a regression model solely based on the extra-linguistics, as shown in Equation~\ref{eq:control}, and then compute the residual error of that model as in Equation~\ref{eq:error}. This error is ultimately used as the outcome label in the final step of RFA.

\noindent{\bf Step 2: Factor selection.} Adaptation to many factors can increase the model parameters drastically. We explore multiple options for selecting a subset of factors from the available extra-linguistic variables. First, we consider manually selected extra-linguistic factors that are known to influence language use more than others. Second, we use the correlation of the factors with the outcome. Last, we use PCA, an unsupervised method to generate new, lower-dimensional factors from the originals. The purpose of factor selection is to reduce the variance and chance of overfitting.

\noindent{\bf Step 3: Factor adaptation.} We modify the original factor adaptation approach to account for the larger number of factors and features in this task. First, we normalize selected factors by min-max scaling and then multiply the language features by these selected factors as shown in Eq.~\ref{eq:adaptation}. As we describe later on in Section~\ref{sec:baselines}, we use n-grams and topics as separate feature sets of language data, so adaptation gives us two corresponding sets of features: adapted n-grams and adapted topics.
We then standardize the adapted features using Z-scores and perform feature reduction on each of the four sets of features separately. These reduced features sets are concatenated into a single large set and fed as input to a learning algorithm in the next step.

\noindent{\bf Step 4: Residual Prediction.} The final step, as shown in Eq.~\ref{eq:rfa}, is to learn the residual errors of the extra-linguistic control model, using language and adapted language features. Here, we first apply feature selection and reduction on each language feature set: topics, n-grams, adapted-topics and adapted-n-grams. Then we put all of them into a single feature space on which we learn a model to predict the residual error from Step 1. To produce the final outcome predictions, the predicted error from this model is combined with the predicted outcomes of the extra-linguistic control model from Step 1.

The choice of feature selection is vital for RFA, both due to the fact that it multiplies the number of language features by the number of extra-linguistic features, and because it uses extra-linguistic features at two levels, one separately and one in integration with language features, potentially leading to overfitting. In Section~\ref{sec:feature-selection}, we investigate different methods for feature selection to improve RFA's performance.

As Fig.~\ref{fig:rfa} shows, RFA is structured similarly to residualized control. However, residualized control uses language data at its final step, whereas RFA uses both language and adapted language, which is obtained using the factor adaptation technique. This helps RFA to benefit from the advantages of both residualized control and factor adaptation. In other words, RFA combines linguistic and extra-linguistic features on both the feature/data level and the model level.
Eq.~\ref{eq:rfa} formulates the RFA method, in which $\epsilon$ comes from Eq.~\ref{eq:error} and $X_{A}$ is defined in Eq.~\ref{eq:adaptation}.

\begin{equation}
    \label{eq:rfa}
    \epsilon \simeq \gamma \times \lbrack X_{L} , X_{A} \rbrack + \lambda
\end{equation}

\section{Evaluation Setup}
Our task is to predict various community-level outcomes based on publicly available data, including social media and other extra-linguistic data such as socioeconomic and demographic information. We focus on two health-related outcomes: heart disease mortality rate~\cite{Eichstaedt2015psychological} and percent fair/poor health life~\cite{culotta2014estimating}; one psychology-related outcome: life satisfaction~\cite{schwartz2013characterizing}; and two economy outcomes: Increased real estate price rate and foreclosure rate~\cite{zamani2017using}. Our high-level approach is to train separate regression models for each outcome. For each county, the input is a set of tweets posted by users from that county as well as aggregate values of socioeconomic and demographic variables for the county, including median income, percentage with bachelors degrees and median age. The full list of socioeconomic/demographic variables are given in Section \ref{sec:data}. The open-source Differential Language Analysis ToolKit was used for the entire analysis pipeline (feature extraction through modeling)~\cite{DLATKemnlp2017}\footnote{Available at \url{https://github.com/dlatk}}. 

\begin{table*}[ht!]
    \begin{center}
        \resizebox{\textwidth}{!}{
            \begin{tabular}{lll*{11}{c}}
            \toprule
                 \textbf{Domain} & & \textbf{Lang.} & \multicolumn{5}{c}{\textbf{3 Socio-Demographic Factors}} & \multicolumn{5}{c}{\textbf{All Factors}} \\
                 
                  \cmidrule(l{5pt}r{5pt}){4-8} \cmidrule(l{5pt}r{5pt}){9-13}
                 & & & \textit{Controls} & \textit{Added-} & \multirow{2}{*}{\textit{RC}} & \multirow{2}{*}{\textit{FA}} & \multirow{2}{*}{\textit{RFA}} & \textit{Controls} & \textit{Added-} & \multirow{2}{*}{\textit{RC}} & \multirow{2}{*}{\textit{FA}} & \multirow{2}{*}{\textit{RFA}} \\
                 
                &  & & \textit{Only} & \textit{Controls} &  &  &  & \textit{Only} & \textit{Controls} &  & &  \\
                \midrule
            Health & HD &  \ColorCell{0.585}{.423}{.655}{.62} & \ColorCell{0.423}{.423}{.655}{.62} & \ColorCell{0.590}{.423}{.655}{.62} & \ColorCell{0.620}{.423}{.655}{.62} & \ColorCell{0.628}{.423}{.655}{.62} & \ColorCell{0.638}{.423}{.655}{.62} & \ColorCell{0.515}{.423}{.655}{.62} & \ColorCell{0.597}{.423}{.655}{.62} & \ColorCell{0.630}{.423}{.655}{.62} & \ColorCell{0.636}{.423}{.655}{.62} &  \textbf{\ColorCell{0.657}{.423}{.655}{.62}}* \\
            & FP &  \ColorCell{0.602}{.434}{.682}{.632} & \ColorCell{0.434}{.434}{.682}{.632} & \ColorCell{0.606}{.434}{.682}{.632} & \ColorCell{0.619}{.434}{.682}{.632} & \ColorCell{0.647}{.434}{.682}{.632} & \ColorCell{0.647}{.434}{.682}{.632} & \ColorCell{0.609}{.434}{.682}{.632} & \ColorCell{0.632}{.434}{.682}{.632} & \ColorCell{0.657}{.434}{.682}{.632} & \ColorCell{0.685}{.434}{.682}{.632} & \ColorCell{0.680}{.434}{.682}{.632} \\
            \hline
            Psych. & LS & \ColorCell{0.214}{0.148}{0.396}{0.326} & \ColorCell{0.148}{0.148}{0.396}{0.326} & \ColorCell{0.219}{0.148}{0.396}{0.326} & \ColorCell{0.292}{0.148}{0.396}{0.326} & \ColorCell{0.308}{0.148}{0.396}{0.326} & \ColorCell{0.338}{0.148}{0.396}{0.326} & \ColorCell{0.326}{0.148}{0.396}{0.326} & \ColorCell{0.352}{0.148}{0.396}{0.326} & \ColorCell{0.376}{0.148}{0.396}{0.326} & \ColorCell{0.353}{0.148}{0.396}{0.326} & \ColorCell{0.396}{0.148}{0.396}{0.326}* \\
            \hline
            Econ. & IP & \ColorCell{0.245}{0.072}{0.402}{0.266} & \ColorCell{0.072}{0.072}{0.402}{0.266} & \ColorCell{0.243}{0.072}{0.402}{0.266} & \ColorCell{0.266}{0.072}{0.402}{0.266} & \ColorCell{0.274}{0.072}{0.402}{0.266} & \ColorCell{0.307}{0.072}{0.402}{0.266} & \ColorCell{0.240}{0.072}{0.402}{0.266} & \ColorCell{0.226}{0.072}{0.402}{0.266} & \ColorCell{0.330}{0.072}{0.402}{0.266} & \ColorCell{0.344}{0.072}{0.402}{0.266} & \textbf{\ColorCell{0.402}{0.072}{0.402}{0.266}}* \\
            
            & FC & \ColorCell{0.153}{0.128}{0.276}{0.197} & \ColorCell{0.128}{0.128}{0.276}{0.197} & \ColorCell{0.156}{0.128}{0.276}{0.197} & \ColorCell{0.197}{0.128}{0.276}{0.197} & \ColorCell{0.218}{0.128}{0.276}{0.197} & \ColorCell{0.238}{0.128}{0.276}{0.197} & \ColorCell{0.160}{0.128}{0.276}{0.197} & \ColorCell{0.161}{0.128}{0.276}{0.197} & \ColorCell{0.209}{0.128}{0.276}{0.197} & \ColorCell{0.240}{0.128}{0.276}{0.197} & \textbf{\ColorCell{0.276}{0.128}{0.276}{0.197}}* \\
            \hline
            \hline
            
            & Avg. & \ColorCell{0.360}{0.241}{0.482}{0.398} & \ColorCell{0.241}{.241}{.482}{.398} & \ColorCell{0.362}{.241}{.482}{.398} & \ColorCell{0.398}{.241}{.482}{.398} & \ColorCell{0.415}{.241}{.482}{.398} & \ColorCell{0.434}{.241}{.482}{.398} & \ColorCell{0.370}{.241}{.482}{.398} & \ColorCell{0.394}{.241}{.482}{.398} & \ColorCell{0.440}{.241}{.482}{.398} & \ColorCell{0.452}{.241}{.482}{.398} & \textbf{\ColorCell{0.482}{.241}{.482}{.398}}* \\
            \bottomrule
             \end{tabular}
        }
        \caption{$R^2$ (variance explained) of residualized factor adaptation (RFA) versus baseline models. Results are shown for 3 hand-picked factors (age, race, education) as well as all factors. RC is residualized control and FA is factor adaptation. Each row is color-coded separately, from red (lowest value) to green (highest values). Bold and * indicate a significant ($p < .05$) reduction in error over the next best model (bold) and FA (*), respectively, according to paired t-tests. }
        \label{tbl:3vsall}
    \end{center}
\end{table*}

\begin{table*}[ht!]
    \begin{center}
        \small
            \begin{tabular}{lll*{11}{c}}
            \toprule
                 & & \multirow{2}{*}{\textit{Lang.}} &  \textit{Controls} & \textit{Added-} & \multirow{2}{*}{\textit{RC}} & \multirow{2}{*}{\textit{FA}} & \multirow{2}{*}{\textit{RFA}} \\
                 
                &  & & \textit{Only} & \textit{Controls} &  &  &    \\
                \midrule
            Health & HD & \ColorCell{0.765}{0.718}{0.811}{0.78}  & \ColorCell{0.718}{0.718}{0.811}{0.78} & \ColorCell{0.773}{0.718}{0.811}{0.78} & \ColorCell{0.794}{0.718}{0.812}{0.78} & \ColorCell{0.798}{0.717189}{0.811}{0.78} &  \textbf{\ColorCell{0.811}{0.718}{0.811}{0.78}}* \\
            & FP & \ColorCell{0.776}{0.776}{0.828}{.8} &  \ColorCell{0.781}{0.776}{0.828}{.8} & \ColorCell{0.795}{0.776}{0.828}{.8} & \ColorCell{0.811}{0.776}{0.828}{.8} & \ColorCell{0.828}{0.776}{0.828}{.8} & \ColorCell{0.825}{0.776}{0.828}{.8} \\
            \hline
            Psych. & LS & \ColorCell{0.463}{0.466}{0.63}{0.604} & \ColorCell{0.571}{0.466}{0.63}{0.604} & \ColorCell{0.594}{0.466}{0.63}{0.604} & \ColorCell{0.614}{0.466}{0.63}{0.604} & \ColorCell{0.595}{0.466}{0.63}{0.604} & \ColorCell{0.630}{0.466}{0.63}{0.604}* \\
            \hline
            Econ. & IP & \ColorCell{0.496}{0.488}{0.633}{0.53} &  \ColorCell{0.490}{0.488}{0.633}{0.53} & \ColorCell{0.476}{0.488}{0.633}{0.53} & \ColorCell{0.575}{0.488}{0.633}{0.53} & \ColorCell{0.587}{0.488}{0.633}{0.53} & \textbf{\ColorCell{0.634}{0.488}{0.633}{0.53}}* \\
            
            & FC & \ColorCell{0.391}{0.396}{0.527}{0.43} &  \ColorCell{0.401}{0.396}{0.527}{0.43} & \ColorCell{0.401}{0.396}{0.527}{0.43} & \ColorCell{0.457}{0.396}{0.527}{0.43} & \ColorCell{0.490}{0.396}{0.527}{0.43} & \textbf{\ColorCell{0.526}{0.396}{0.527}{0.43}}* \\
            \hline
            \hline
            
            & Avg. & \ColorCell{0.578}{0.578}{0.685}{0.63} & \ColorCell{0.592}{.578}{.685}{.63} & \ColorCell{0.608}{.578}{.685}{.63} & \ColorCell{0.650}{.578}{.685}{.63} &  
            \ColorCell{0.659}{.578}{.685}{.63} & \textbf{\ColorCell{0.685}{.578}{.685}{.63}}* \\
            \bottomrule
             \end{tabular}
        \caption{Pearson-$r$ of residualized factor adaptation (RFA) versus baseline models (for comparison to other work which uses Pearson-$r$ as the accuracy metric). Results are only shown for all factors. RC is residualized control and FA is factor adaptation. Each row is color-coded separately, from 
        red (lowest value) to green (highest values). Bold and * indicate a significant ($p < .05$) reduction in error over the next best model (bold) and over FA (*), respectively, according to paired t-tests. }
        \label{tbl:pearsonr_all}
    \end{center}
\end{table*}

\subsection{Data Set} 
\label{sec:data}

Our evaluation dataset includes information from three sources: (1) \textit{language data} from Twitter messages, (2) \textit{extra-linguistic data} consisting of 11 socioeconomic and demographic variables, and (3) \textit{outcome data} consisting of 5 county-wise outcomes from 3 categories: Health, Psychology, and Economy.

Our language data can be divided into two groups, (1) for Health and Psychological outcomes and (2) for Economical outcomes.
The language data we use for Health- and Psychology-related outcomes was derived from Twitter's 10\% random stream collected from July 2009 to February 2015 and includes 1.64 billion tweets~\cite{giorgi2018}\footnote{Available at \url{https://github.com/wwbp}}. 
For Economy outcomes, we used the language data from \cite{zamani2017using}. This data was derived from Twitter's $1\%$ random stream collected from 2011 to 2013 and includes 131 million tweets. In both cases, the tweets were mapped to counties based on users' self-reported location strings using the procedure proposed by \cite{schwartz2013characterizing}.

The extra-linguistic data consists of 11 variables used in previous work: 4 \textit{socioeconomic} variables including median income, unemployment rate, percentage of bachelors degrees, and percentage of high school degree, as well as 7  \textit{demographic} variables including median age; percentage: female, black, Hispanic, foreign-born, married; and population density~\cite{census2010ussencus}. 
All variables were obtained from the US Census~\cite{census2010ussencus}, and we henceforth refer to them collectively as \textit{extra-linguistic features}. This dataset is only collected every 10 years, so the 2010 US Census is the most recent dataset for all of the \textit{socioeconomic} and \textit{demographic} variables at the county level.

We consider 5  county-wise measurements as outcomes, 2 health-related (heart disease mortality rate, fair/poor health life), 1 psychological (life satisfaction), and 2 economic (yearly increased real estate price rate, yearly foreclosure rate). Health and psychological data was gathered from the Centers for Disease Control and Prevention (2010b)~\nocite{cdc2010} and contains between 1,630 to 1,749 counties, depending on the outcome. The economic outcomes, which have been used previously in \cite{zamani2017using}, were gathered for the year 2013 from Zillow\footnote{http://www.zillow.com/research/data/}. They contain 427 counties' foreclosure rate and 717 counties' increased real estate price rate. 

\subsection{Baselines} \label{sec:baselines}

Our baselines consist of a controls-only prediction model and a language-only prediction model. 

\noindent{\bf Controls-only.} The controls-only model is a simple regression model trained over all the 11 extra-linguistic features. 

\noindent{\bf Language-only.} Building this baseline consists of three main steps:
extracting linguistic features, performing feature reduction, and running ridge-regression~\cite{goeman:2016}. 
Our linguistic features are n-gram features (1-3 grams) and topic features which include mentions of 2,000 LDA~\cite{blei2003latent} derived topics previously estimated from social media~\cite{schwartz2013personality}.

For language data, we first pruned the sparse n-gram features to only include those that were mentioned in at lease a percentage of the counties, then due to the importance of word count in performance of language predictive models\cite{zamani2018predicting} we exploit a word count threshold and drop counties with fewer words. Then we run a correlation threshold to only keep the highest correlated features and finally we perform a randomized principal components analysis (RPCA), an approximate PCA based on stochastic re-sampling~\cite{rokhlin2009arandomized}. We apply the correlation threshold and RPCA steps for n-grams and topics independently.

For language data associated with health and psychology outcomes, we pruned the sparse n-gram features to only include those that were mentioned in at least 95\% of the counties, and used 20,000 as the word count threshold, resulting in 27,250 n-grams total.

With only 1,749 training instances (one per county), feature selection and dimensionality reduction become necessary for avoiding overfitting. We first limit the features to the top 10,000 n-grams with the highest linear relationships to each outcome. As the topic features are more informative than a single n-gram, we choose to retain all 2,000 topics at this step. Then after performing RPCA we only keep 100 features for each group of ngrams and topcs. 

For the language data associated with economy outcomes, we pruned the n-gram features to only include those that were mentioned in at least 10\% of the counties, and used 10,000 as the word-count threshold, resulting in 8,897 n-grams across 717 training instances. We use the top 8,000 n-grams and the top 1,500 topic features with the highest linear relationships to each outcome. at the end by applying RPCA we limit the dimension of each feature set to 100.

We compare performances of residualized control, factor adaptation and residualized factor adaptation (RFA). For all these models, we use the same settings as above to generate language features.

\section{Results}

\subsection{Comparison of RC, FA, and RFA} 
We first compare factor adaptation (FA), residualized control (RC), and residualized factor adaptation (RFA) using three manually selected factors: age, race (percentage of black population), and education (percentage with bachelor's degree) rates. These three factors are often used as ``controls'' in prior work~\cite{schwartz2013characterizing,culotta2014estimating,Eichstaedt2015psychological,curtis}\footnote{Income has also been used frequently but it has been shown to correlate strongly with education rates.} and also represent examples of demographic and socioeconomic measurements. 

In order to ensure a fair comparison, we use the same extra-linguistic features for all models. As mentioned earlier, a naive method is to directly combine the extra-linguistic features with language ones in a single feature set. Here, we also compare this simple model, which we call \textit{added-controls}, with the other three models. In addition, we consider a linear model solely using extra-linguistics, which we call \textit{controls only}.
Evaluation is done using 10-fold cross-validation. $R^2$, or variance explained, is used to measure accuracy.

Table~\ref{tbl:3vsall} compares results in terms of variance explained, when using the three hand-picked factors vs. using all 11 extra-linguistic factors (Since past work has also used the Pearson-$r$ metric, Table~\ref{tbl:pearsonr_all} shows the same results for all factors in terms of Pearson-$r$). As the table shows, FA outperforms controls only, added-controls, and residualized control. RFA does even better and outperforms FA on both the hand-picked factors and when using the entire set of factors. These results demonstrate the complementary nature of the residualized control and factor adaptation approaches and the benefits of combining them.

Even though adding controls directly, as in the ``added-controls'' column, works better than language-only and controls-only models, it is worse than any other model that exploits both language and extra-linguistic data. 
This motivates the need for combining different types of features in both an additive (residualized control) and multiplicative (factor adaptation) style. 

Overall, these results show the power of RFA over the other models. RFA's improvement over FA was statistically significant for 4 out of 5 outcomes, and 3 out of 5 for residualized control. Recall that added-controls, residualized control, FA, and RFA all have access to the same set of information. The gains of RFA over FA show that RFA's structure utilizing residualized control is better suited for combining extra-linguistic and language-only features.

\begin{table}[t]
\centering
\small
\begin{tabular}{c|cccc}
 & \begin{tabular}[c]{@{}c@{}}No\\ FS\end{tabular} & \begin{tabular}[c]{@{}c@{}}Separated\\ FS\end{tabular} & \begin{tabular}[c]{@{}c@{}}Combined\\ FS\end{tabular} & \begin{tabular}[c]{@{}c@{}}Early\\ FS\end{tabular} \\
 \hline
HD                   & 0.656                                           & \textbf{0.657}                                         & 0.65                                                  & 0.639                                               \\
FP                   & 0.678                                           & \textbf{0.68}                                          & 0.676                                                 & 0.661                                               \\
LS                   & 0.364                                           & 0.396                                                  & 0.391                                                 & \textbf{0.401}                                      \\
IP                   & \textbf{0.425}                                  & 0.402                                                  & 0.392                                                 & 0.336                                               \\
FC                   & 0.187                                           & \textbf{0.276}                                        & 0.268                                                 & 0.241                                               \\
\hline
AVG                  & 0.462                                           & \textbf{0.482}                                         & 0.475                                                 & 0.456                                              
\end{tabular}
\caption{Comparing $R^2$ using different methods of feature selection. Outcomes are heart disease (HD), fair/poor health (FP), life satisfaction (LS), increased price (IP), and foreclosure rate (FC). FS stands for feature selection. Bold cells have the highest $R^2$ for each outcome.}
\label{table:featureSelection}
\end{table}

\subsection{Feature Selection} \label{sec:feature-selection}
Here we investigate the impact of feature selection on the overall performance of RFA. We consider three different combination of adaptation and feature selection, as well as adaptation without any feature selection: (1) \textit{SeparatedFS:} apply feature selection separately on language features and adapted language features; (2) \textit{CombinedFS:} combine language features and adapted language features into one feature set and then apply feature selection; (3) \textit{EarlyFS:} apply feature selection on language features, then apply adaptation on the selected features; and (4) \textit{NoFS:} perform adaptation without any feature selection. Table~\ref{table:featureSelection} shows the performance of each method on all 5 outcomes, as well as the the average. SeparatedFS performs better than the others in 3 out of 5 cases, as well as leading the average $R^2$ across all 5 outcomes. In addition, it produces the most stable results in comparison to the other methods. We therefore use this method for RFA.

\begin{figure}[t!]
    \centering
    \includegraphics[width = .9\columnwidth]{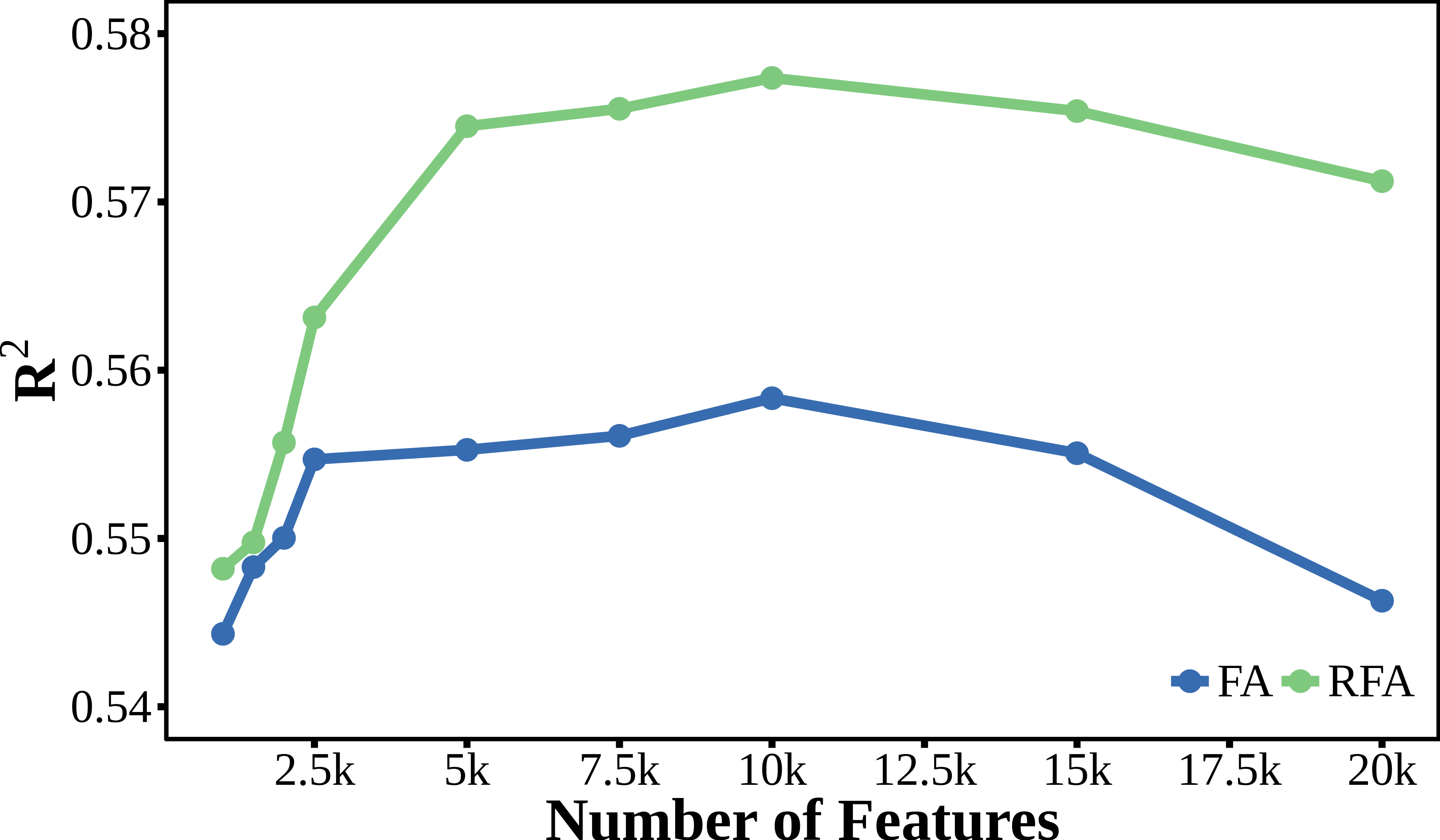}
    \caption{Effect of increasing the number of selected features in univariate feature selection  on both factor adaptation (FA) and residualized factor adaptation (RFA) by looking at the average $R^2$ among health and psychological outcomes. All 11 factors are used in all cases.}
    \label{fig:k_best}
\end{figure}

We perform another experiment to find the best parameter for the univariate feature selection, that is, the value of $k$ when selecting the $k$-best n-gram features. 
Figure~\ref{fig:k_best} shows the results of varying the number of features used for FA and RFA. We report the average $R^2$ across the 3 health- and psychology-related outcomes. In general, selecting more features leads to better results, though eventually performance does begin to suffer. Recall that our feature selection approach is to first select the $k$-best n-gram features based on their linear relationship with the outcome, then do a PCA on these $k$ features to obtain a reduced-dimension vector. Even though the feature selection doesn't directly increase the size of our models, it effectively increases the amount of information available to the models, leading to the positive trends we see in Figure~\ref{fig:k_best}.

\subsection{Increasing Factors and Factor Selection}
This experiment has two objectives: first to find out how the number of factors affects performance, and second to find an automated way to select a good subset from the extra-linguistic factors. Here we vary the number of factors from 1 to 11 (i.e. all factors) and compare the effects on RFA, FA, and RC. 
Factor selection in this experiment is done in two ways, supervised and unsupervised. For the supervised selection we use Recursive Feature Elimination, in which for each $k$, the least significant factors are recursively dropped until only $k$ factors remain. For unsupervised selection, we use PCA to build $k$ new factors with the highest variance. 

\begin{figure}
    \centering
    \includegraphics[width=.48\columnwidth]{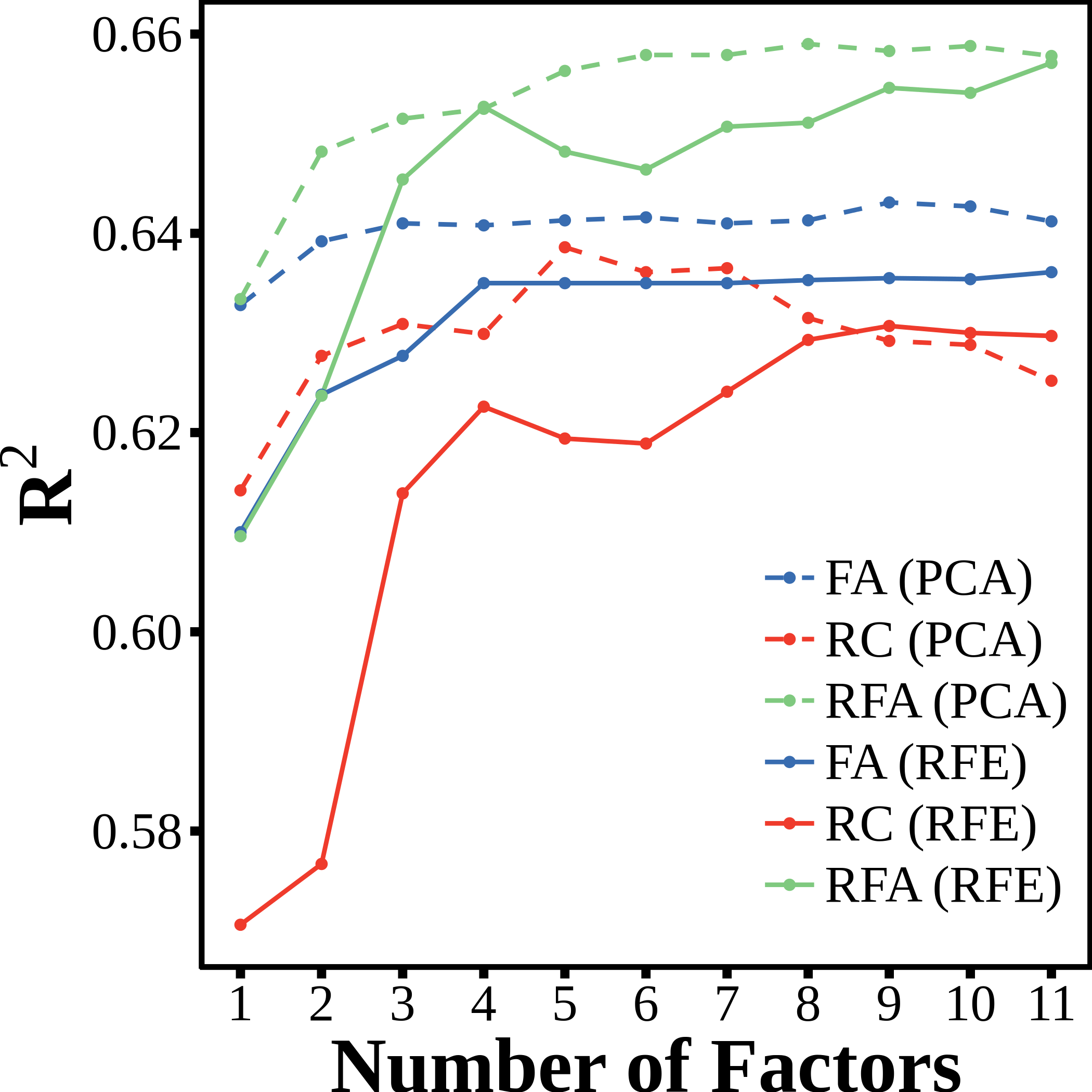}
    \includegraphics[width=.48\columnwidth]{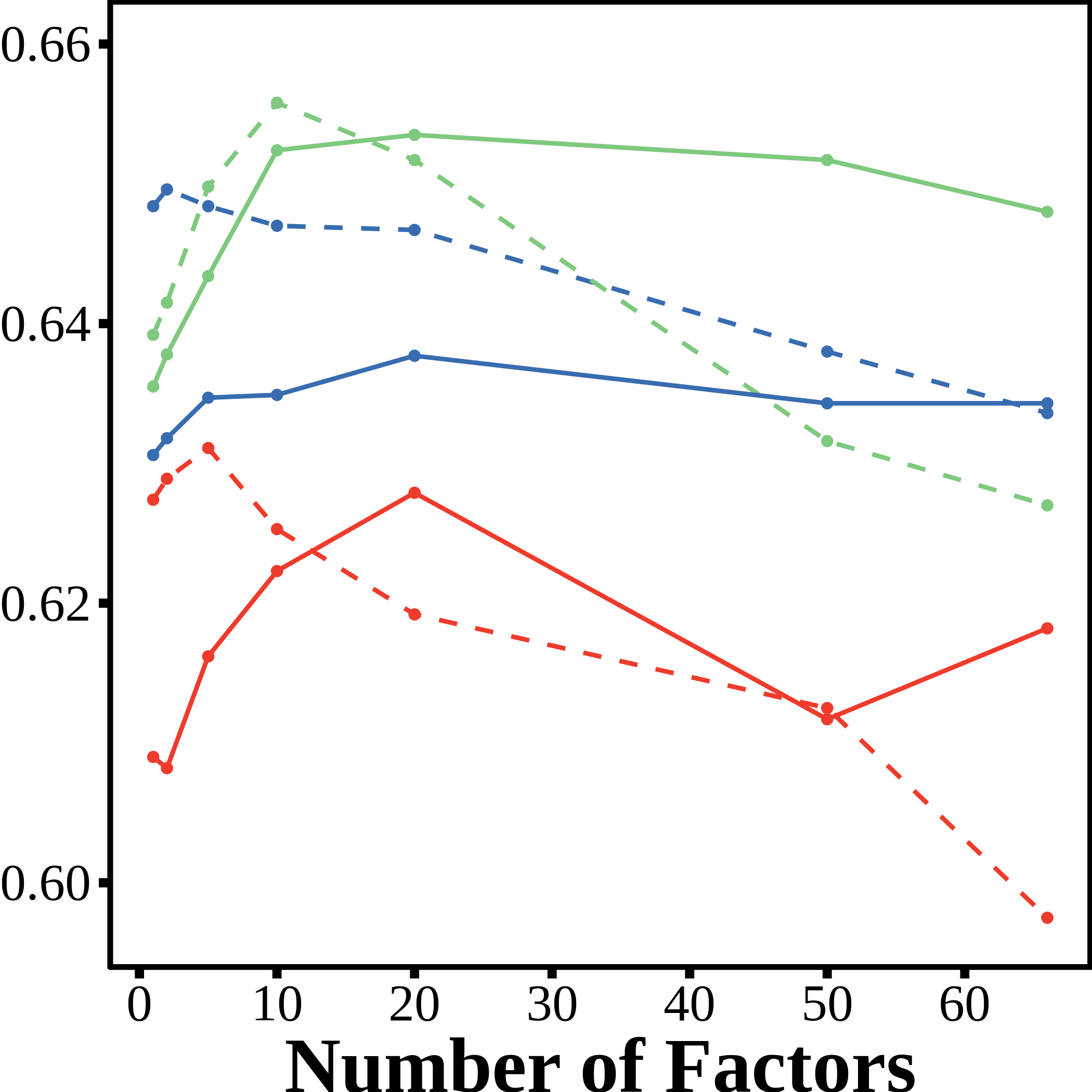}
    \caption{Effect of increasing number of factors on $R^2$ of residualized factor adaptation (RFA), factor adaptation (FA) and residualized controls (RC) for heart disease outcome. Factors are obtained through Recursive Feature Elimination (RFE) or PCA. Left plot is with original factors, and right plot is with interaction factors (the product from pairing factors). 
    }
    \label{fig:factorsSelectionHea}
\end{figure}

The left of Figure~\ref{fig:factorsSelectionHea} shows how the performance of RFA, FA and RC change for heart disease outcome as the number of factors increase, using both PCA and RFE as factor selection methods.

RFA outperforms FA at every factor number, and begins to outperform RC as the number of factors increases. RFA's performance, in general, tends to increase as we add more factors. Using PCA, RFA reaches close to its best performance very quickly, requiring only 5 or 6 factors; adding more factors results in longer runtimes for minimal gain. However, in the case of RFE, using more factors appears to be worthwhile. FA and RC both quickly plateau, or even decline, as more features are added.

Since performance generally improved as more factors were added, we explored adding more factors beyond the 11 that are available to us. To this end we create new factors by multiplying the existing factors with one another. 
To account for variance in the factor ranges, we first min-max normalize each factor. Then we consider every pair of factors and multiply their normalized values together and re-normalize these new values to create a new factor. This gives a total of $55$ new factors in addition to the original 11. We rerun our experiments with this new pool of 66 factors.

The right of Figure~\ref{fig:factorsSelectionHea} shows the results of using this expanded pool of factors. Here, the performance begins to taper off beyond 15 factors for both FA and RFA.
Overall, PCA obtains its best performance with only a few factors, but then begins to suffer as more factors are added. RFE, on the other hand, tends to perform worse than PCA initially but remains relatively stable as more factors are added.
These newly-created features turned out to be less effective than the original eleven, suggesting that the trade-off in increasing factors via combination is not worthwhile.

Overall, even though reducing the number of factors through PCA-based factor selection could not beat the best accuracy, it is still very competitive. Given the potentially huge number of features obtained through factor adaptation, this slight decrease in performance may be worth potential increases in runtime. RFE-based factor selection, however, helps with neither the runtime nor the performance.

\section{Conclusions}

Language-based prediction tasks involving communities can benefit from both socio-demographic factors and linguistic features. Because this information comes from different sources and has different distributions, effective mechanisms are needed for combining them. In this paper, we present \textit{residualized factor adaptation}, a method that unites two ways of approaching this problem, one where strong community attributes are augmented (i.e.~additive use of factors) with weak but noisy language features, and the other where the contextual differences in language use are mediated via community attributes (i.e. adaptation to community factors). The proposed method effectively combines the complementary benefits of both residualized control and factor adaptation approaches to yield substantial gains over differing community-level prediction tasks across three domains. 
We see this work as part of a growing need for application-oriented approaches that not only leverage large data effectively by themselves, but do so in the context of other social scientific information that is already available and valuable.

\section*{Acknowledgments}
This work was supported, in part, by a grant from the Templeton Religion Trust (ID $\#$TRT0048). The funders had no role in study design, data collection and analysis, decision to publish, or preparation of the manuscript.

\bibliographystyle{acl_natbib_nourl}

\end{document}